\documentclass{article}

\usepackage{arxiv}

\usepackage[utf8]{inputenc} 
\usepackage[T1]{fontenc}    
\usepackage{hyperref}       
\usepackage{url}            
\usepackage{booktabs}       
\usepackage{amsfonts}       
\usepackage{nicefrac}       
\usepackage{microtype}      
\usepackage{lipsum}
\usepackage{graphicx}
\usepackage[ruled, vlined]{algorithm2e}
\usepackage{amsmath}
\usepackage{subfig}
\usepackage{xcolor}

\title{Evolving Indoor Navigational Strategies Using Gated Recurrent Units In NEAT}

\author{
  James Butterworth \\
  Department of Computer Science\\
  University of Liverpool\\
  Liverpool, L69 3BX \\
  \texttt{j.butterworth2@liverpool.ac.uk} \\
   \And
  Rahul Savani \\
  Department of Computer Science\\
  University of Liverpool\\
  Liverpool, L69 3BX \\
  \texttt{rahul.savani@liverpool.ac.uk} \\
  \And
  Karl Tuyls \\
  Department of Computer Science\\
  University of Liverpool\\
  Liverpool, L69 3BX \\
  \texttt{k.tuyls@liverpool.ac.uk} \\
}

\begin{document}
\maketitle

\begin{abstract}
Simultaneous Localisation and Mapping (SLAM) algorithms are expensive to run on smaller robotic platforms such as Micro-Aerial Vehicles. Bug algorithms are an alternative that use relatively little processing power, and avoid high memory consumption by not building an explicit map of the environment. Bug Algorithms achieve relatively good performance in simulated and robotic maze solving domains. However, because they are hand-designed, a natural question is whether they are globally optimal control policies. In this work we explore the performance of Neuroevolution - specifically NEAT - at evolving control policies for simulated differential drive robots carrying out generalised maze navigation. We extend NEAT to include Gated Recurrent Units (GRUs) to help deal with long term dependencies. We show that both NEAT and our NEAT-GRU can repeatably generate controllers that outperform I-Bug (an algorithm particularly well-suited for use in real robots) on a test set of 209 indoor maze like environments. We show that NEAT-GRU is superior to NEAT in this task but also that out of the 2 systems, only NEAT-GRU can continuously evolve successful controllers for a much harder task in which no bearing information about the target is provided to the agent.
\end{abstract}


\section{Introduction}

Smaller robotic platforms such as Miro-Aerial Vehicles (MAVs) have the potential to carry out tasks in indoor environments that are often too dangerous for a human to do. Some of these tasks include: search-and-rescue after natural disasters, radioactivity monitoring, and the surveillance of safety critical infrastructure. The size and manoeuvrability of these systems allows them to access areas that would be inaccessible for larger robots. In order to perform useful tasks in these domains, the robotic systems need to be equipped with a number of specific algorithms such as: maximal coverage, collision avoidance, and navigation.


Autonomous robotic navigation is an important task to optimise for many robotic systems operating autonomously in an unknown environment. Simultaneous Localisation and Mapping (SLAM) is the process whereby an agent constructs a map of the environment whilst also localising itself within that environment. Once a map of the environment has been generated, the agent can perform path planning through the map to some desired goal. The success of this planning is dependent on the success of SLAM. 


SLAM itself is a thoroughly developed field that has robust methods for dealing with static, structured and limited size environments \cite{handbook_robotics}. This has been demonstrated by its success on a number of different robotic platforms including BigDog \cite{bigdog}, Unmanned-Aerial Vehicles (UAVs) \cite{range, droneslam}, helicopters \cite{Thrun2006} and even autonomous vehicles \cite{bresson}. Despite these successes, SLAM is still a relatively computationally expensive algorithm which can be problematic for smaller robotic platforms with limited computational power such as smaller MAVs.


An alternative suite of algorithms known as Bug Algorithms aim to navigate through a maze without building an explicit representation of the map, this is a lot less computationally expensive and memory intensive. The main idea behind these algorithms is that the agents have limited sensor capabilities and react locally to objects such as walls when they come into contact with them. Often, the agent knows the relative position (or the azimuth angle) of the goal but is not aware of the structure of the environment. A number of these algorithms have been developed for simulation and simple robotic applications however there is not one individual algorithm that can solve all environments and it is often the case that each algorithm has its own subset of environments that it performs well in.


Even though these Bug Algorithms are quite successful, they are all hand designed. This raises the question as to whether there exists more efficient and effective algorithms or control policies for these environments that have not yet been conceived of. The recent surge of interest in using Machine Learning methods for optimising agent controllers has lead to success in a wide variety domains with many breakthroughs coming from the areas of Reinforcement Learning (RL) and Neuroevolution (NE). Both of these techniques automatically discover control policies for agents situated in an environment, however the method of optimisation differs considerably. Both methods have been shown to outperform humans at a number of different tasks with one of the most widely used benchmarks being the Atari domain \cite{Mnih2015, atari_2, DeepNeuro}. This suggests that human level skill or policies of an equivalent skill are by no means always a global optima in the solution space and it is often the case that an undiscovered, better performing policy exists waiting to be found.

In this work we use the Neuroevolutionary technique known as Neuroevolution of Augmenting Topologies (NEAT) \cite{NEAT}. NEAT evolves a network structure as well as the weights, which often leads to much smaller networks than those in which the structure has been hand designed. This can significantly reduce the computational power and memory load which is especially advantageous for Evolutionary Robotics. Furthermore, we choose NE over RL due to the fact that it is conceptually simpler to evolve controllers for continuous action spaces than the RL equivalent algorithms and NE does not suffer the same divergence issues often seen in policy gradient methods.


In this work, we explore the application of Neuroevolution to the domain of indoor navigation via simulated Foot-bots - a differential drive robot commonly used in swarm robotic research \cite{footbot}. We aim to answer the following questions:

\begin{enumerate}
    \item Can Neuroevolutionary techniques (specifically NEAT) generate controllers that perform better than a specific type of bug algorithm called I-Bug, thereby showing that hand designed bug algorithms are sub optimal?
    \item Can controllers that outperform I-Bug be evolved repeatably and robustly?
    \item Does the inclusion of long term memory units, specifically Gated Recurrent Units (GRUs) \cite{gru}, into NEAT lead to better performances on these tasks?
\end{enumerate}

We show that both NEAT and NEAT-GRU can generate controllers that outperform I-Bug on a test set of 209 indoor maze like environments and we show that successful controllers can be evolved repeatably. We also show that there is a small advantage of using NEAT-GRU in this task. One limitation of I-Bug is its reliance on a bearing sensor that provides the agent with its relative bearing to the target. A natural question is whether evolution can produce solutions in a domain in which bearing information is not provided. We show that NEAT-GRU consistently evolves solutions that can solve this harder task where as normal NEAT could not solve this task at all. This suggests that NEAT-GRU has the ability to produce solutions that perform significant cognition and are able to accumulate state information over time in order to locate the target. 

Our main contributions can be summarised as follows:

\begin{enumerate}
    \item We introduce NEAT-GRU, an extension to the neuroevolutionary technique NEAT. Within NEAT-GRU, GRUs can be mutated into NEAT networks just like hidden nodes and their parameters are optimised via mutation operators.
    \item We show empirically that NEAT-GRU can produce control policies that outperform I-Bug on a large test set of navigation domains, providing concrete evidence that hand designed bug algorithms are sub optimal.
    \item We show that these superior control policies can be evolved repeatably.
    \item We provide evidence that suggests NEAT-GRU is superior than NEAT at producing solutions for these navigation domains thereby inferring that long term memory units provide additional assistance to the control networks.
    \item We show that NEAT-GRU can continuously produce solutions that solve a much harder navigation task in which no bearing information about the target is provided to the agent. NEAT was unable to produce any solutions for this task, suggesting the need for more complex memory structures when evolving control policies for more complex, real world domains.
\end{enumerate}

In Section \ref{related_work} we introduce the reader to the I-Bug algorithm and discuss some of the other related maze solving results in the areas of evolutionary computation and reinforcement learning. Section \ref{methodology} briefly introduces NEAT, GRUs and describes our NEAT-GRU system. Section \ref{experimental_setup} explains how the experiments are setup. Section \ref{results} highlights the results in which we find empirical evidence suggesting the advantages of our NEAT-GRU in 2 different domains. Finally, we finish with a discussion of the results and consider avenues for future work in Section \ref{discussion}.

\section{Related Work} \label{related_work}

\subsection{Bug Algorithms} \label{bug_algorithms}

The most simple maze navigational algorithm is wall following. By continuously aligning to the wall
on either the left or right of the agent, an exit is guaranteed to be found as long as the maze has no disjoint walls or loops, this is known as a simply connected maze. Apart from the long run-time of this algorithm, it is also unsuitable for indoor navigation as indoor environments are often not simply connected - they contain loops or disjoint sections which can cause the wall following to get stuck in an infinite loop.

\begin{figure}[h!]
  \caption{The `Com' bug algorithm. The agent moves along a straight line towards the target until an obstacle is met, it will then follow the obstacle until it can continue on its path to the target.}
  \centering
    \includegraphics[scale=0.4]{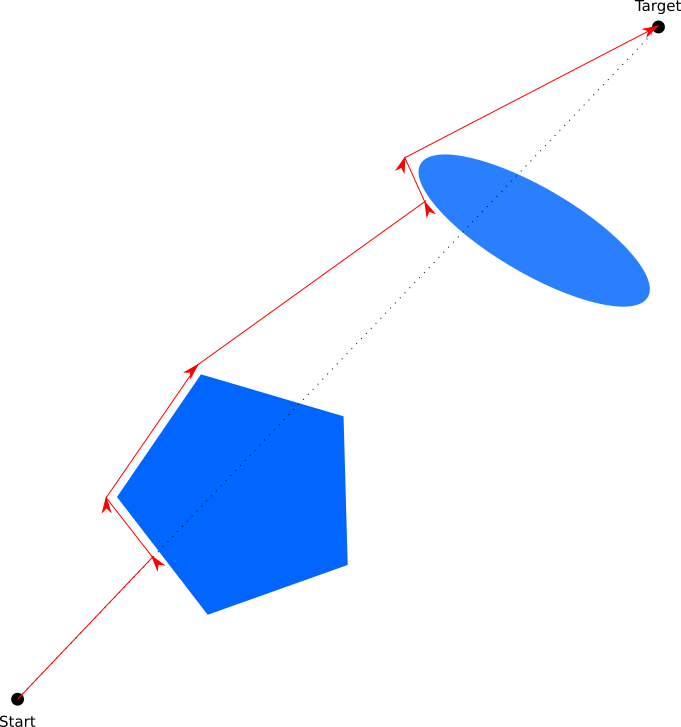}
    \label{fig:com}
\end{figure}

Bug Algorithms do not require simply connected environments and can deal with unknown obstacles or arbitrary shapes. Lumelsky and Stepanov \cite{com} pioneered these algorithms by introducing three: Com, Bug1 and Bug2. The simplest of the three, Com (Figure \ref{fig:com}), is carried out as follows:

\begin{enumerate}
  \item Move along a straight line towards the target until one of the following occurs:
  \begin{enumerate}
    \item The target is reached.
    \item An obstacle is met. Follow the obstacle boundary in the prespecified local direction (i.e. left). Go to step 2.
  \end{enumerate}
  \item Leave the obstacle boundary at a point $z$ if the agent can move along a direct line towards the target. Go to step 1.
\end{enumerate}

Although Com works successfully in a number of environments, there are a number of cases in which the agent can get stuck in an infinite loop. Bug1 and Bug2 are more complex versions of Com which aim to overcome some of the issues such as infinite loops and long path lengths. 

A number of other bug algorithms have been developed \cite{alg1, alg2, vis1, vis2, tanbug, cautious, insert, distbug}, each with their own advantages and disadvantages. Most note worthy for the application of these algorithms to robotics is I-Bug \cite{ibug}. In I-Bug, the target is assumed to have a wireless intensity beacon which continuously provides the distance to the target. It is also assumed that the agent can detect when it is horizontally aligned with the target and has some form of short range proximity sensors that can detect walls. These constraints are most appropriate for use on MAVs due to the availability of similar physical sensors that can determine these values. Previous work has demonstrated collision avoidance on MAVs via use of Ultra-Wideband Frequency chips which are able to communicate intra-drone distances to one another \cite{basti}. There is also no requirement for I-Bug to have a global coordinate system, whereas this is a requirement for most other bug algorithms but can often be problematic to implement in real robotic scenarios. 

In the following description of I-Bug, `intensity' refers to the intensity of a wireless signal with respect to the target, it is assumed to be a maximum of 1 when the agent is at the target. I-Bug consists of 3 possible actions or movement primitives:

\begin{itemize}
    \item $u_{fwd}:$ The robot goes straight forward in the direction it is facing, stopping only if: 1) it contacts an obstacle, 2) hits the target, 3) detects a local maximum of intensity along its line of motion.
    \item $u_{ori}:$ The robot rotates counterclockwise, stopping only when it is aligned with the target.
    \item $u_{fol}:$ The robot travels around an obstacle boundary counterclockwise, maintaining contact to its left at all times, stopping only when it reaches a local maximum intensity.
\end{itemize}

With this combination of sensors and actions, I-Bug is able to carry out Algorithm 1. In this algorithm $h_i(x)$ refers to the intensity of the wireless signal at the agents' current state, $x$. This algorithm stores two values throughout its execution: $i_{L}$ and $i_{H}$. The intensity $i_{H}$ is the intensity recorded when the agent contacts an obstacle following the termination of $u_{fwd}$ and $i_{L}$ is the value recorded just prior to the execution of $u_{fwd}$. Conceptually, they are the intensities at which the agent \textit{hits} and \textit{leaves} objects. $i_{L}$ is used to determine whether the agent has moved following the $u_{fwd}$ action and $i_{H}$ is used to determine when to terminate wall following and continue to move towards the target. Wall following will only terminate if a local maximum of intensity is reached (this is part of the movement primitive) and then conditioned on whether the current intensity is greater than $i_{H}$ - in other words, it will only stop wall following when it was approaching the target and has now begun to move away from it (local maximum) then conditioned on whether it is closer to the target than it was when it first contacted the obstacle.

\begin{algorithm}
\SetAlgoLined
\SetKwRepeat{Do}{do}{while}
\While{not at target}{
  $i_{L} \leftarrow h_{i}(x)$\;
  Apply $u_{ori}$\;
  Apply $u_{fwd}$\;
  \If{$h_{i}(x)=1$}{
  $at\_target \leftarrow true$\;
  \textbf{terminate}\;
  }
  \If{$i_{L} \neq h_{i}(x)$}{
  $i_{H} \leftarrow h_{i}(x)$
  }
  \Do{$h_{i}(x) \leq i_{H}$}{$u_{fol}$}
  }
\caption{I-Bug}
\label{ibug_alg}
\end{algorithm}


Further analysis of the I-Bug algorithm and other bug algorithms highlights that certain variables - in the case of I-Bug: $i_{H}$ - must be stored over long time periods at certain times. Therefore it is very possible that controllers optimised to carry out policies of similar or greater performance than I-Bug would require mechanisms that support the ability to remember information over long term periods. This is one of the main motivations for including GRUs into NEAT.


\subsection{Evolutionary Methods}

Evolutionary methods optimise controllers via the operations of selection, crossover and mutation as oppose to, for example, gradient descent in deep RL approaches. Given the recent success of non-gradient based Genetic Algorithms on the Atari framework - a benchmark that was previously dominated by Deep RL algorithms - an argument can be made that following a gradient is not always the best approach to achieve globally optimal policies \cite{DeepNeuro}.  

A comparative study of generalised maze solvers explores the performance of controllers evolved via NEAT using objective based search, novelty search (NS) and a hybrid of the two \cite{Shorten2015}. Novelty Search is an evolutionary algorithm that searches the space of \textit{behaviours} rather than trying to optimise an explicit objective. It does so by assigning a Behaviour Characteristic (BC) to each individual and then gives a reward based upon how different the BC of the individual is with respect to the current population \textit{and} an archive of previously novel individuals. This encourages exploration into novel behavioural spaces which can help mitigate deception - a problem whereby explicit objectives do not illuminate a path to the global optima. The aim of \cite{Shorten2015} was to evolve simulated robot controllers to solve unseen mazes as oppose to learning a policy to solve the same maze. It was found that both NS and the hybrid approach solved significantly more mazes than the objective based approach and were able to generalise to larger and more difficult mazes. The advantages of NS over objective search are echoed in the earlier work on NS \cite{NS1, NS3, NS2} and also in Quality Diversity algorithms (a hybrid of NS and objective based methods) \cite{QD} in which maze domains are a key benchmark task. 

A common benchmark task used to test a neuroevolutionary algorithms' memory capability is the T-Maze domain. This domain has a number of forms: one being a discrete state and action space and the other being of a more continuous nature, there also exists the double T-Maze domain that contains more reward locations than the original maze. The main requirement for success in all these tasks is the ability to remember the location of a large reward over a significant number of time steps. It is often the case that standard recurrent connections struggle in this domain as they are unable to deal with long term dependencies.

A discrete version of the T-maze and double T-maze domains were solved by using neuromodulated plasticity whereby the synaptic weights of normal neural network connections were modified online via hebbian rules \textit{and} modulatory neurons \cite{Soltoggio2008}. It was shown that the networks that were evolved using fixed connections performed significantly worse on both domains whereas networks that were evolved with modulatory neurons outperformed both fixed weight and Hebbian architectures in the double T-maze domain achieving a maximum score. Other work \cite{Risi2009, Risi2010NS} shows the advantages of NS in evolving similar neuromodulatory networks on the discrete version of the T-maze domain and \cite{Ollion2012} argues the reason is that evolving memory is highly deceptive and proposes a behavioural diversity technique similar to NS that achieves similar performance gains. An alternative way to encourage individual neurons to exhibit specific memory functions in the T-maze domain is by evolving networks \textit{and} neurons in two separate populations, which helps due to neurons of different sub functions being prevented from mating \cite{Gomez2005}.

Indirect encoded versions of NEAT have also been tested on these T-maze domains via the Adaptive HyperNEAT \cite{Risi2010} and Adaptive ES-HyperNEAT \cite{Risi2012} algorithms. The main idea behind the original HyperNEAT algorithm \cite{HyperNEAT} is that the weights of the synaptic connections are determined by querying a Compositional Pattern Producing Network (CPPN) with the 2 dimensional coordinates of the connection being queried. Given that the connection strength is a function of its position, the CPPN is topologically aware and also has the ability to produce repeating motifs or symmetries similar to the human brain. Extending HyperNEAT to include connections that are modified online \cite{Risi2010, Risi2012} generates networks that can solve the continuous T-maze domain by instilling them with memory. 

The Evolvable Neural Turing Machine (ENTM) is an algorithmically simpler version of the original Neural Turing Machine (NTM) in that it can be trained using evolutionary operators and is not required to be differentiable \cite{ENTM1, ENTM2, ENTM3}. It has been shown in \cite{ENTM1} that an ENTM can be trained to solve the continuous T-maze domain and the \textit{continuous} double T-maze domain - a task that had not yet been solved by any other algorithm so far. Also worth of note is \cite{GRUMB} in which a Gated Recurrent Unit with a Memory Block is introduced. In this architecture, each GRU has an associated memory block which it can explicitly read and write to - similar to the ENTM - with the idea being that the memory block is `shielded' from irrelevant information. Even though this work does not technically conduct experiments with an agent situated in a T-maze, one of the experiments carried out is a Sequence Recall task which is analagous to the discrete T-maze experiment. It is shown that this new GRU memory block architecture significantly outperforms previous NEAT architectures with Long Term Memory Cells.

Another interesting solution to the discrete versions of the T-maze and double T-maze tasks is by evolution of Continuous Time Recurrent Neural Networks (CTRNNs) \cite{CTRNN}. The state of the neurons in a CTRNN are described by a set of differential equations with respect to time. The idea being that the weights of the synapses are kept constant throughout the run but the internal network dynamics facilitates long term memory via its dynamic neuron potentials.

Although not benchmarked on the T-maze domains, Minimal Criteria evolutionary techniques are effective in solving other maze domains \cite{MCNS, MCC}. These techniques work by allowing all solutions that meet some minimal criteria into the reproductive gene pool, which helps to greatly improve the diversity of the population. Interestingly, in \cite{MCC}, minimal criteria methods are used to coevolve both the agents and mazes, leading to an increasingly complex pool of environments as well as solutions to these environments.

\subsection{Reinforcement Learning}

Autonomous maze solving has been extensively explored within the RL community with the most recent work being able to learn navigation policies from raw visual input \cite{Tessler, Kempka, Kulkarni, robot_3, Tai, Mirowski, Jaderberg, Mnih, Oh}. Many of these algorithms can learn to navigate in complicated mazes with the assistance of Long Short Term Memory units (LSTM) \cite{Mirowski, Jaderberg, Bakker, Mnih} which allows the agents to store relevant information for a long period of time. The advantages of using LSTMs over feedforward or pure recurrent architectures in navigation domains has been repeatedly shown \cite{Mirowski, Bakker, Mnih}. Despite there being a small number of examples of RL algorithms (mainly policy gradient variants) approaching navigation using a continuous action space \cite{Pfeiffer}, most of the current approaches use a discrete action space.

\subsection{Novelty Of Our Work}

T-maze domains are very effective at evaluating the abilities of evolved solutions at keeping track of long term dependencies, however they are not aimed at evolving generalised maze solving agents. Agents or robots operating in a lifelike domain will come across many types of navigational challenges that were not part of their training environment. We are more interested in agents that can evolve generalised behaviours and to this end, training on a T-maze domain would not instill our agents with the behavioural requirements of interest to us. However, the previous work on the T-Maze domains does highlight to us the importance of using memory components capable of retaining long term information in maze domains and motivated the use of specific long term memory units in our own work.

The work most similar to ours and the only work that addresses generalised maze solving is \cite{Shorten2015}. However, the comparison is based upon one metric: the number of mazes solved. In our work, we additionally consider the distance it takes each agent to solve the maze, we believe this to be an equally important metric to consider. Furthermore, \cite{Shorten2015} does not consider the use of long term memory components whereas we augment our version of NEAT to mutate in GRUs.

Our domain has a fully continuous state and action space, whereas the action space in the continuous T-maze domain in \cite{ENTM1} consists of 3 discrete action choices which arguably makes for easier control compared to our continuous wheel speed range. 

Finally, in a lot of the work on NS and in the work on generalised maze solving \cite{Shorten2015}, the agent has four pie-slice sensors that informs the agent of the direction towards the goal. Similarly, in some of the T-maze domains the agent is given some signal at the beginning of the run that informs it of the direction in which to turn in order to find the reward. We evaluate our NEAT-GRU on a task in which the agent does not have access to \textit{any} information regarding the direction of the goal and therefore has to determine this based on distance information alone.

\section{Methodology} \label{methodology}

The evolutionary algorithm used within this work is NEAT \cite{NEAT}, a widely used Neuroevolutionary algorithm that evolves a networks' structure as well as its connection weights. It does this by starting minimally (without any hidden nodes) and complexifying via mutations that insert new nodes and links into the evolving networks. NEAT manages to foster more structurally complex solutions via a process of speciation. The speciation mechanism works by assigning networks with significantly different structure to new species that only compete amongst themselves. This helps protect new structural mutations from the competitive full population whilst they are trying to optimise around the new structure. It is often the case that this new structure can lead to a significant increase in performance but will initially perform poorly until the network has learned to utilise it.

In this work we modify NEAT to include Gated Recurrent Units (GRUs) in a new system called NEAT-GRU. GRUs were chosen as oppose to LSTMs due to a similar performance in other domains despite having to optimise a smaller number of parameters \cite{grucompare}. In this sense NEAT-GRU is very similar to NEAT-LSTM introduced in \cite{rawal} however there are less parameters to optimise and our NEAT-GRU is tested in a continuous control domain. 

A GRU unit is a recurrent unit designed to store information over long periods of time. It does this by applying a number of operations to the previous hidden state of the unit. There are 2 main operations or `gates': the reset gate and the update gate. The reset gate determines how to combine the new inputs with the previous hidden state and subsequently provides a new candidate value to store. The update gate decides how much of the previous hidden state information should be kept hold of and how much of the new candidate information to store instead. This provides a mechanism by which the current hidden state can be kept around indefinitely as long as the update gate chooses not to change it. Or, given the input `conditions' it may determine that it is appropriate to store the new information instead.

The GRU units in NEAT-GRU are inserted into the networks in the same way as any other hidden nodes, they are mutated in with some probability. The parameters or weights of the GRUs are modified in the same way as the other NEAT weights, via a perturbation of the original weight where the perturbation is drawn from a uniform distribution between the negative and positive of a mutation power variable. Furthermore, just as in NEAT, there is occasionally a severe mutation in which the perturbation value completely replaces the weight as oppose to just being appended to it. For simplicity, crossover is not used in NEAT-GRU.

\section{Experimental Setup} \label{experimental_setup}

In the following section we describe the setup of 3 experiments: I-Bug on the generalised maze solving task with bearing, evolution on the generalised maze solving task with bearing and finally evolution on the no bearing task with a simplified environment. I-Bug is not tested on the no bearing experiment due to the fact that it cannot operate without a bearing sensor.

The robotic simulator ARGoS is used in this work for both the baseline I-Bug experiments and for the training and testing of the evolved solutions \cite{argos}. ARGoS has been designed to be as accurate of a representation of the real world as possible providing detailed models of commonly used robots in academia. In this work we use the provided Foot-Bot model \cite{footbot}, due to the fact that it is equipped with 24 local laser sensors, a range and a bearing sensor, which are the same sensors that are required for the I-Bug algorithm. 

\subsection{I-Bug}

\begin{figure}[h!]
  \caption{An example of one of the randomly generated environments used in the test set. The environment has a number of deceptive rooms and corridor structures. One of the robots is the target and is motionless throughout the run whereas the other robot contains the navigation algorithm and aims to find the other robot.}
  \centering
    \includegraphics[scale=0.23]{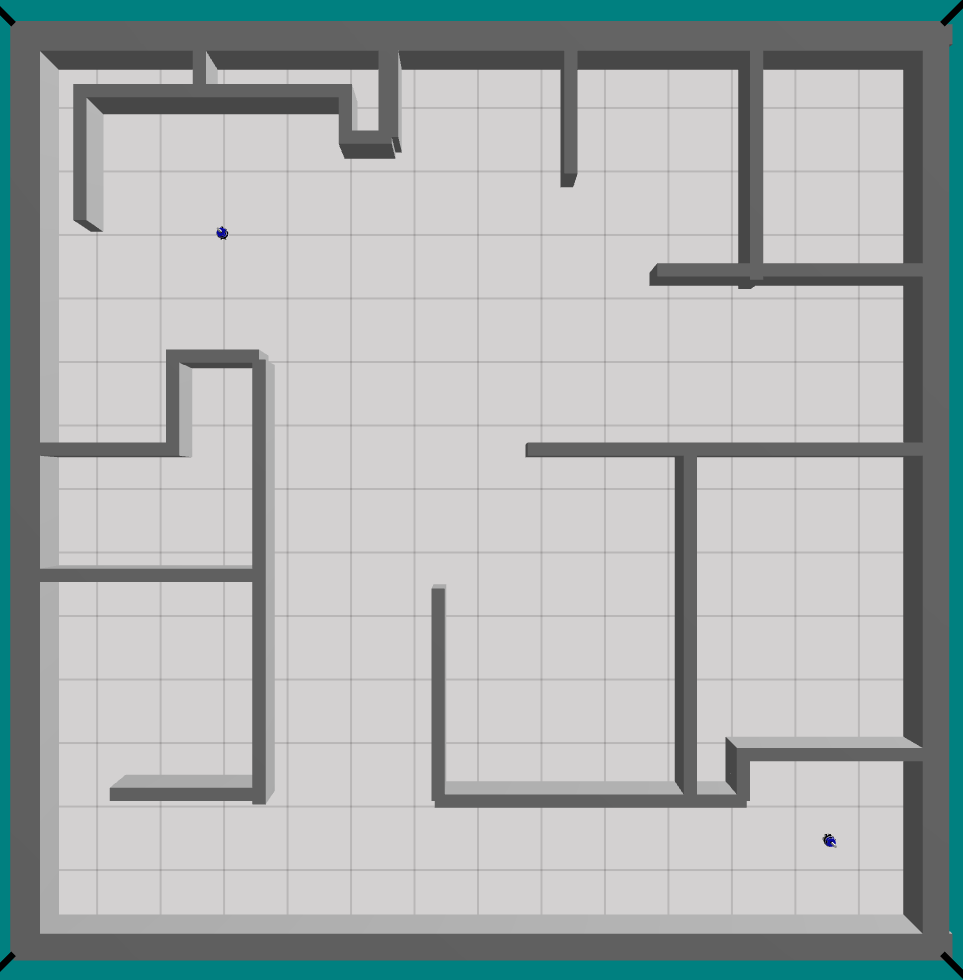}
    \label{fig:env_example}
\end{figure}

In order to get a good idea of the performance of I-Bug a simulated Foot-Bot carrying out the algorithm was evaluated on 209 randomly generated test environments which were used as a test set throughout this work. These environments were the same test set that was originally used in \cite{kim_bug}: a comparative survey of Bug Algorithms that also uses ARGoS and the same Foot-Bot setup as our own work. The maze generation algorithm used in \cite{kim_bug} leads to a wide variety of environments that closely resemble real life indoor office or living environments with corridor and room like structures. An example of one of these test environments is shown in Figure \ref{fig:env_example}.

The following specification of simulation configuration is the same as in \cite{kim_bug}. For the I-Bug evaluations the positions of the 24 proximity sensors on the Foot-Bot were modified in the following way. 20 of the sensors were positioned in a wedge at the front of the robot to simulate a depth sensor or stereo camera for obstacle detection. There are also 2 sensors at 90 degrees to the front of the robot and 1 sensor directly behind the robot. The size of the sensors was also increased to two meters. This was modified due to the fact that a hand designed wall following behaviour was more robust with a higher density of sensors at the front of the robot. A wall following behaviour based upon this sensor structure was used as the $u_{fol}$ motion primitive described in Section \ref{bug_algorithms}. The rest of the I-Bug algorithm was implemented exactly as in Algorithm \ref{ibug_alg}.

The performance of I-Bug was then tested on the 209 test set environments with a constant size of $14m \times 14m$. The agent had 300 simulated seconds to navigate through each environment - this time limit seemed sufficient to allow an agent to navigate to a number of dead ends, realise its mistake and go down an alternative route to the target. The success percentage of I-Bug was recorded as the number of mazes in which the target was found as a percentage of the total number of mazes in the test set. Furthermore, the agents' trajectory lengths were recorded and normalised by divided through by the A* path length, which represents the shortest possible path through the maze. This A* length is calculated using a grid connectivity graph approach over a grid of size $140 \times 140$ representing the environment. This grid resolution results in a sufficiently accurate path length.

\subsection{Evolved Solutions}


\subsubsection{Bearing Sensor} \label{evolved_solutions_bearing_sensor}

In order to encourage a generalised maze solving behaviour the environments used in training are randomly generated according to the same parameters that generated the test set. Each genome is tested on the same set of 10 mazes each generation, but for each generation this set of 10 mazes is newly generated. The entire evolutionary run is ran for 1000 generations with a population size of 150. Each agent has 300 simulated seconds to find the target. Every 25 generations, the 3 best genomes from the current generation, the best 2 genomes from previous generation and the best genome from two generations ago are tested on the test set and their performance is recorded. Due to the time taken to test genomes on the test set, only a small number can be tested. This genome collection procedure was chosen based upon the idea that the highest performing genomes on the training set will most likely (but not always) have the highest performance on the test set. Also, genomes from previous generations are chosen due to the fact that the procedurally generated training set of the current generation might be a set of environments that, by chance, does not give an accurate representation of the test set. 20 runs are performed using both NEAT and NEAT-GRU.

Given that evolution can quickly generate a wall following behaviour without having to have a dense sensor wedge at the front of the agent or requiring long proximity sensors, the Foot-Bot sensors were altered for the evolutionary experiments. Instead of 24 sensors, only 12 are used and they are situated equally spaced around the robot; furthermore, the size of the sensors was reduced to $0.2m$. Using less sensors reduces the size of the initial networks by reducing the number of network inputs. The network inputs are the 12 proximity sensors, 1 range sensor giving the distance to the target and 2 bearing sensors: one is the relative bearing of the agent to the target going clockwise from the front of the agent and the other going counter-clockwise from the front of the agent. The network outputs are the speeds of the left and right wheels of the Foot-Bot.

In order to evolve solutions according to 2 metrics: the number of mazes solved and the trajectory length per A* length, a fitness function had to be designed to optimise both of these values. Initially a weighted sum of both values was used however the evolutionary results only generated policies that maximised only one of the two metrics despite any configuration of weightings assigned to each. The agents would either evolve a wall following behaviour (maximising the number of mazes solved but doing so with a relatively large path length) or evolve a greedy behaviour in which it headed straight for the target (reducing path length but not getting to many targets, often due to the agent getting stuck in rooms).

However, it was noticed that a very small number of policies in the first generation were capable of finding at least one of the 10 targets in the training set. This observation lead to the fitness function, $f_{1}$, used for the evolutionary runs:

\begin{equation}
 f_{1} =
  \begin{cases} 
    \frac{1}{l^{0.5}} & \text{if } maze\_solved \\
    0 & otherwise
  \end{cases}
  \label{eq1}
\end{equation}

where $l$ is the trajectory length per A* length taken to find the target. This fitness function resulted in the best trade off between the two metrics due to the fact that the agent must find the target and then once it had it is rewarded with the inverse of the length taken to get there. An exponent of $0.5$ was chosen in order to `flatten' the function slightly such that a genome was not rewarded too much for doing well in one particular environment. Furthermore, if the agent crashed into a wall the final fitness was divided by 10 in order to deter the agents from crashing.

Initial experiments revealed that tuning the hyperparameters from the standard NEAT parameters resulted in better performance. Particularly worth of note are the add node and add GRU node mutation rates were $0.005$ and $0.003$ respectively - setting these too high resulted in a considerable amount of bloat. Furthermore a higher survival threshold of $0.55$ was seen to assist performance by reducing selection pressure. The weight mutation power was set to $1.5$ for both normal weights and the GRU node weights.

\subsubsection{No Bearing Sensor} 

A much harder task was designed to test the cognitive abilities of NEAT-GRU further. This task reduces the number of sensors leaving just the distance to target as input. This sensor configuration removes the ability of the agent to know its relative orientation towards the target, therefore finding the target must be done by accumulating distance measurements and performing significant cognition in order to ascertain the direction in which to travel. This sensor configuration commonly occurs in robotics where a distance sensor - such as an Ultra-Wideband Frequency chip - provides \textit{just} distance information with respect to other chips, as in the work of \cite{basti}.

Given that this task is much more difficult, the environment used to train and test the evolved solutions was considerably simpler. An environment of size $10m \times 10m$ was used that contained no obstacles apart from a wall around the perimeter. This environment was difficult enough to demonstrate a considerable performance difference between NEAT and NEAT-GRU. All of the proximity sensors were disabled for this task due to the fact that the agents could learn a wall following behaviour that led them close to the goal without having to learn any complex policies.

The agent begins in one corner of the $10m \times 10m$ square and the target in the opposing corner. Each agent is evaluated on the same environment 5 times however the starting orientation of the agent is different for each evaluation. This is to prevent overfitting to a set orientation in which an agent `memorises' a particular path from a particular starting position to the target. Each agent is given 80 simulated seconds to find the target, this is far less than the bearing sensor experiments due to the fact that the environment is smaller and there are no obstacles to navigate around. Like the bearing experiments the network outputs are the left and right wheel speeds of the Foot-Bot whereas the only network input in these experiments is the target distance. 

The fitness function $f_{1}$ in equation \ref{eq1} was not successful in this experiment due to the fact that none of the initial genomes could get to the target, therefore they would all receive a fitness of zero. In order to provide a significant reward gradient $f_{2}$ was used:

\begin{equation}
 f_{2} = (L - d)^{3}
  \label{eq2}
\end{equation}

where $L$ is the diagonal length of the arena (the maximum distance the agent can be from the target) and $d$ is the final distance between the agent and the target at the end of the run. Also, just like the previous experiments, if an agent crashes, the fitness is divided by 10. It was found that cubing the fitness lead to a larger rate of success by greatly rewarding those runs that performed slightly better.

NEAT and NEAT-GRU were ran 10 times each for 5000 generations per run and a population size of 150. A slightly higher mutation rate for adding nodes and gru nodes was used: 0.006 and 0.006 respectively. Furthermore a weight mutation power of $0.5$ was used in order to reduce the rate of dramatic weight changes within the grus. The more common survival threshold of $0.4$ was used.

\section{Results} \label{results}

\subsection{I-Bug}


On the 209 test environments I-Bug achieved a success percentage of $93.3\%$ meaning that it managed to find the target in the maze $195/209$ times. The mean and median of the trajectory lengths per A* lengths over all the environments were $2.4174$ and $1.69$ respectively. I-Bug was not tested on the no bearing experiments because without the relative angle towards the target I-Bug will not work.


\subsection{Evolved Solutions}

\subsubsection{Bearing Sensor} \label{bearing_solutions}

Out of 20 evolutionary runs for NEAT-GRU, 10 of the runs produced genomes (out of those evaluated on the test set - highlighted in Section \ref{evolved_solutions_bearing_sensor}) that outperformed I-Bug at maze success percentage and in the mean of the trajectory length per A* length values over all the environments. 5 of the aforementioned 10 runs contained more than one genome that outperformed I-Bug however, it was often the case that they were similar in behaviour to previous genomes in the same run so it is not worth noting the total number of genomes found that outperformed I-Bug, the number of runs in which evolution could produce at least one outperforming solution is more significant. Furthermore, out of these 20 runs, 2 of the runs produced genomes that outperformed I-Bug in all 3 metrics: maze success percentage and the mean \textit{and} median of the trajectory length per A* length values over all the environments. Despite $2/20$ being quite a low number of successful number of runs according to all of three metrics, the objective performance in training is optimised according to the \textit{mean} of the trajectory length per A* length as oppose to the median.

\begin{figure}%
    \centering
    \subfloat[All of the solutions outperform I-Bug in terms of the number of finishes and the trajectory means per A* mean metrics.]{{\includegraphics[width=7.5cm]{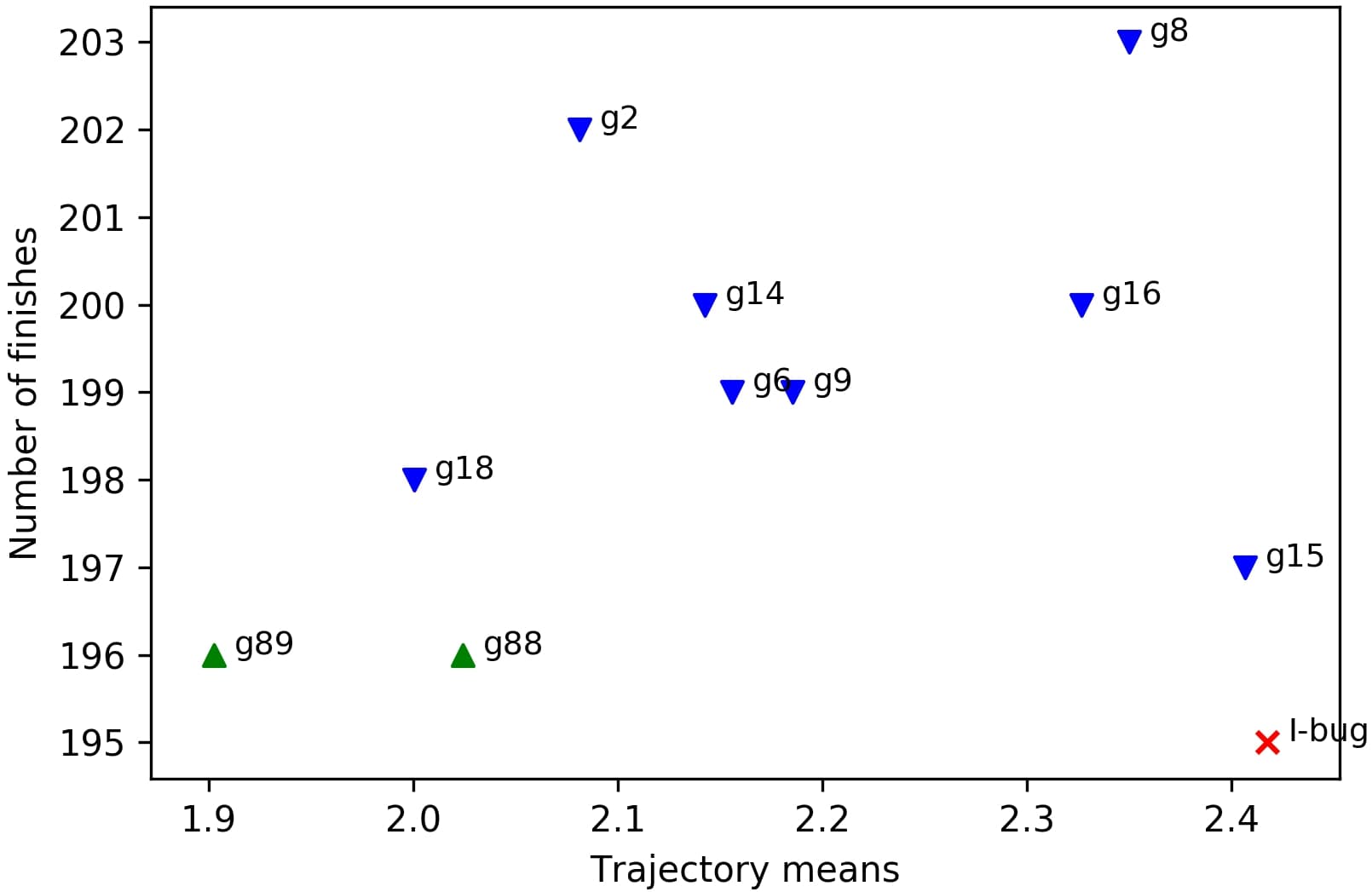} }}%
    \qquad
    \subfloat[I-Bug has a relatively good trajectory median compared to the evolved solutions, with only G88 and G89 outperforming it in all 3 metrics.]{{\includegraphics[width=7.5cm]{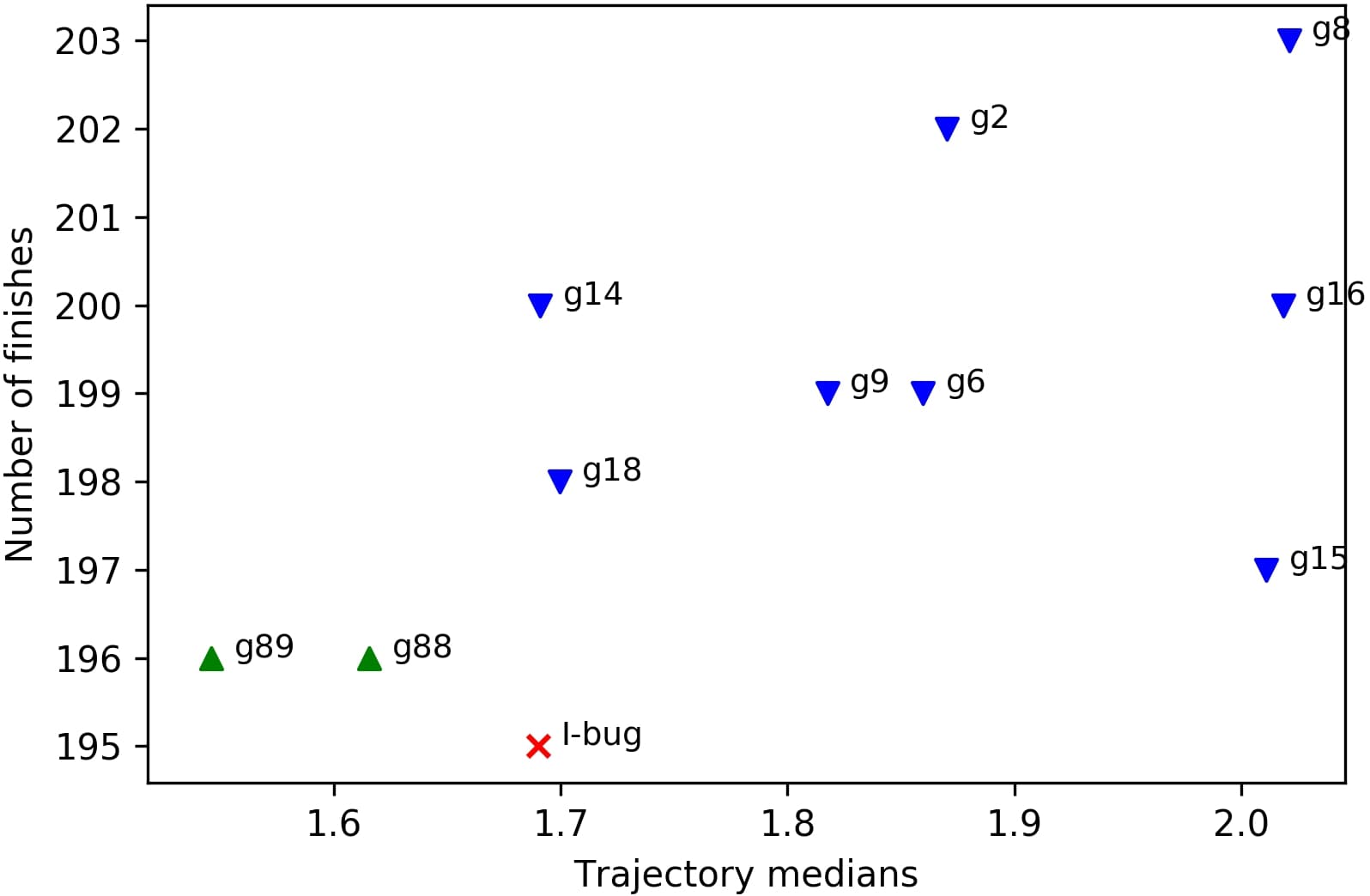} }}%
    \caption{Scatter charts showing the performance metrics for the 10 genomes produced by NEAT-GRU that outperformed I-Bug and I-Bug itself. A smaller trajectory length is more desirable. The 2 solutions that outperformed I-Bug in all 3 metrics are highlighted in green and the 8 solutions that outperformed I-Bug in just 2 metrics are highlighted in blue.}%
    \label{fig:genome_graphs}%
\end{figure}

Figure \ref{fig:genome_graphs} illustrates the performance metrics of the 10 genomes that beat I-Bug in at least 2 metrics. Some of the solutions produced were significantly better than I-Bug (p < 0.0001 based upon trajectory lengths per A* length). For example, one solution named `G89' had a success rate of 196/209 (93.7\%), a trajectory length per a star length mean of 1.9024 and a median of 1.5459. This trajectory length mean is significantly less than the equivalent I-Bug value meaning that the agent was able to get to the target in a much shorter path length despite being able to find more targets than I-Bug. There also exist solutions that have a larger success rate but at the expense of having longer path lengths. For example `G8' had a success rate of 203/209 (97.1\%), a trajectory mean of 2.3499 and a median of 2.0212. It therefore gets to the target on a far larger number of occasions than I-Bug, it does so with a mean trajectory length per A* length less than I-Bug however it does not quite beat I-Bug on the median of this value. A video showing the behaviour of these genomes is available at https://youtu.be/8EqyeuX\_lR0

\begin{figure}
  \caption{A graph showing the average population fitness and the max population fitness during training for both GRU and non-GRU versions of the bearing experiment. It shows the slight fitness increase attributed to the inclusion of GRUs. The results are averaged over 20 runs.}
  \centering
    \includegraphics[scale=0.25]{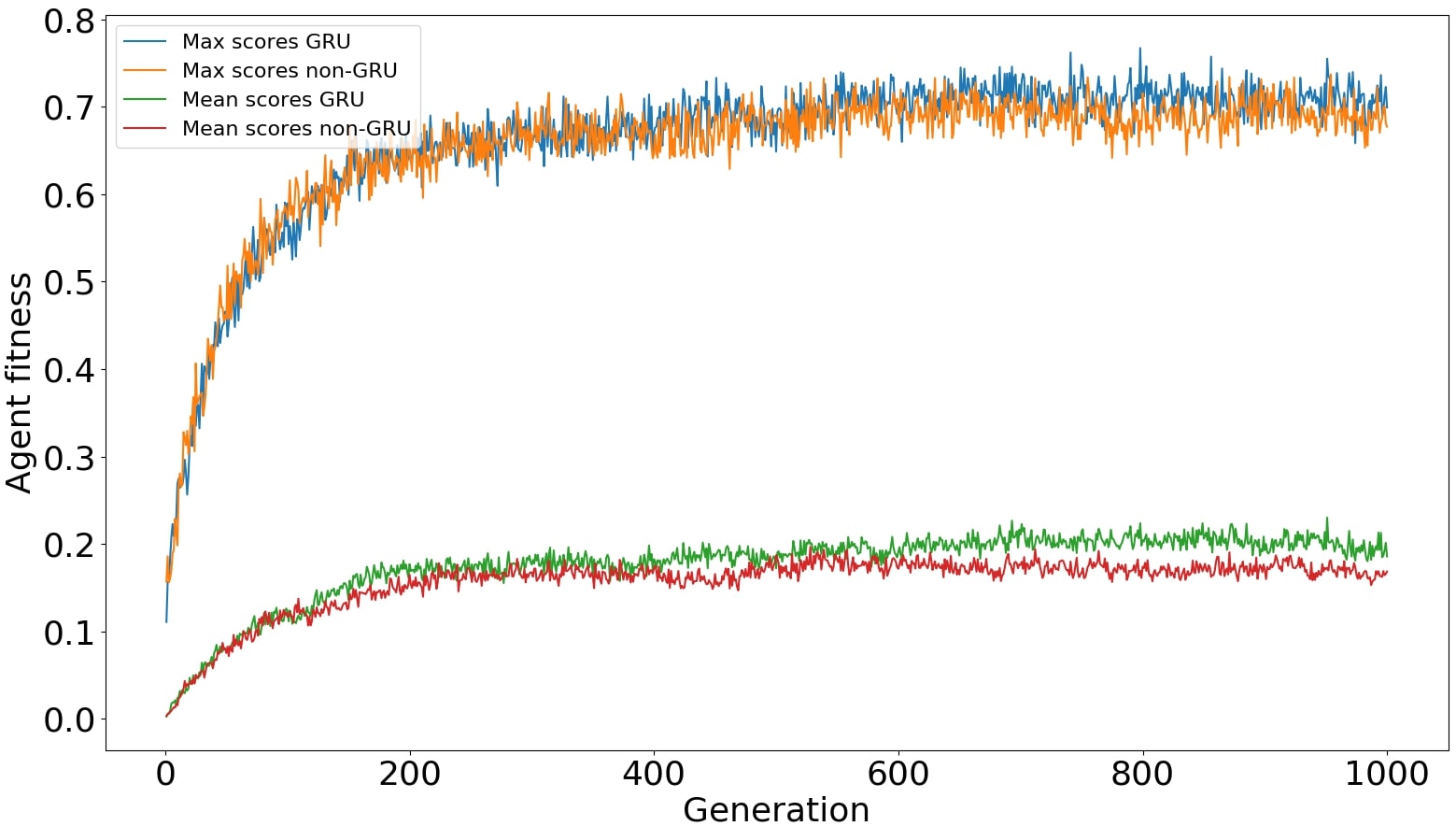}
    \label{fig:train_easy}
\end{figure}

Out of the 20 evolutionary runs for NEAT, 3 of the runs produced genomes that outperformed I-Bug at maze success percentage and in the mean of the trajectory length per A* length values over all the environments. 0 of the winning solutions beat I-Bug in all 3 metrics. These results are highlighted in table \ref{table:bearing_results}. Furthermore, Figure \ref{fig:train_easy} highlights the average and max fitness of the whole population at each generation during training with GRUs and without GRUs. It highlights the slight advantage in performance offered by GRUs.

\begin{table}[h!]
\centering
\begin{tabular}{|l|c|c|}
\hline
         & \multicolumn{1}{l|}{2 metric winners} & \multicolumn{1}{l|}{3 metric winners} \\ \hline
NEAT     & 3/20                                     & 0/20                                     \\ \hline
NEAT-GRU & \textbf{10/20}                            & \textbf{2/20}                            \\ \hline
\end{tabular}
\caption{A table highlighting the number of evolutionary runs out of 20 in which at least one genome outperformed I-Bug on the 209 test environments for both NEAT and NEAT-GRU.}
\label{table:bearing_results}
\end{table}

\subsubsection{No Bearing Sensor}

\begin{figure}
  \caption{A graph showing the maximum fitness so far for the population during training for both GRU and non-GRU versions of the non-bearing experiment. It shows a dramatic fitness increase attributed to the inclusion of GRUs and how the non-GRU version plateaus at a score of 3000. The results are averaged over 10 runs.}
  \centering
    \includegraphics[scale=0.25]{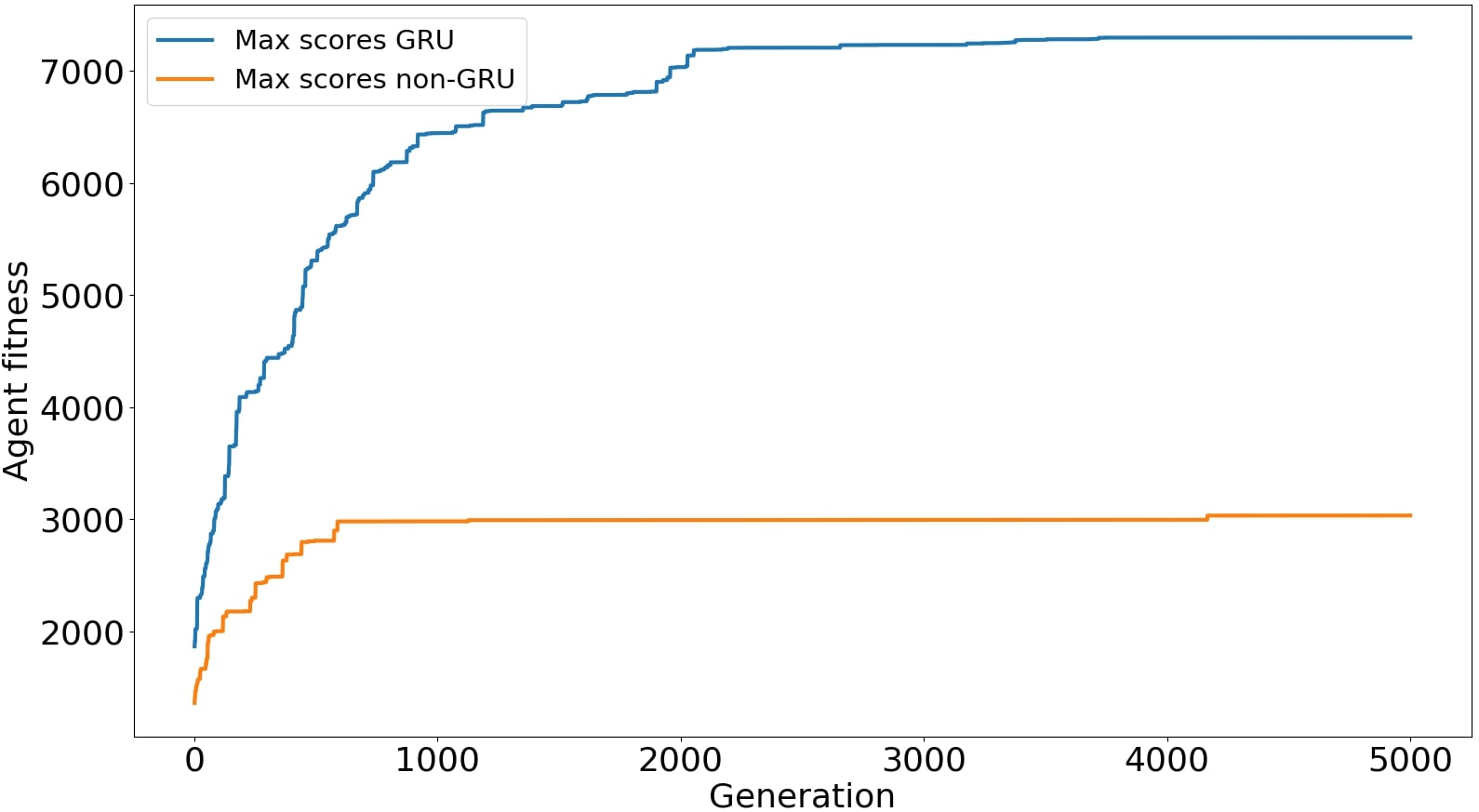}
    \label{fig:train_hard}
\end{figure}

Out of the 10 evolutionary runs for NEAT-GRU, all 10 produced a solution capable of solving the task in all 5 orientations. In contrast, out of the 10 runs using NEAT, 0 of them produced solutions that could solve the task in all 5 orientations. Figure \ref{fig:train_hard} shows the maximum fitness so far for the population during training for the no bearing task. It shows the dramatic increase in performance due to the inclusion of GRUs into the NEAT networks. Source code for all these experiments is available on request.

\section{Discussion \& Future Work} \label{discussion}

The results in section \ref{bearing_solutions} clearly highlight that I-Bug is not a globally optimal solution in the domain of generalised, reactive maze solving. It also shows that evolution and more specifically, NEAT and NEAT-GRU, have the ability to produce better control policies than those that have been previously hand designed. This continues to add to the previous body of work that demonstrates the ability of machine learning techniques to outperform human designed algorithms. Even though the solutions evolved in this work are an improvement on previous bug algorithms, there is no guarantee (and is highly unlikely) that this work has found globally optimal solutions either. We would therefore like to explore the application of Novelty Search in these domains and test whether this method of optimisation can bring further benefits.

This work does highlight that the inclusion of long term memory units into NEAT leads to a better performance in the bearing experiments and is seemingly essential for completion of the no bearing experiments. This fits in with the growing evidence suggesting the inclusion of long term memory units into networks improves performance on maze like tasks. However, what was interesting was that NEAT without GRUs was able to outperform I-Bug, despite networks of this type being unable to store information over long periods. It was also observed during the experiments that networks without any hidden units performed quite well, with one genome achieving a success rate of 189/209 (90.4\%) and a mean path length per A* path length of 2.1. This suggests that the maze following environments used to test I-Bug and other bug algorithms - as well as many real indoor environments - do not actually require much cognition to solve well. 

The results of the no bearing tasks suggests the high relative difficulty of this task in relation to the bearing version. With only the distance to the target at each point in time, the agent does not know in what way to turn in order to approach it unless it builds some form of target location model through time as it accumulates distances. Furthermore, the agent also has to be aware of its own actions in order to build this model because it needs to know how the distances change as a function of its own actions. NEAT theoretically has the ability to do this via recurrent connections that can originate at output nodes however the inclusion of GRU nodes into this greatly helps performance by allowing appropriate information to accumulate. It is also quite surprising that this ability can be learned using a continuous state and action space.

In future, we would like to extend the work on the no bearing experiments further to try and incorporate obstacles back into larger environments and hopefully get comparable performance to other bug algorithms with access to more sensor information. We would then like to extend this into a robotic domain such as with MAVs where the available sensor information would be the target distance and small proximity sensors. Given a swarm of MAVs equipped with these sensors, they could learn to coordinate with respect to each other and navigate to other members of the swarm in complicated indoor environments.

It would also be interesting to analyse how one could optimise GRUs via evolutionary operators in a more efficient manner. Combine different types of mutation operators and explore whether meaningful crossover procedures exist that are beneficial to their optimisation. These ideas will also be explored in future work.

\section{Conclusion} \label{conclusion}

We investigated the ability of evolution, specifically NEAT to evolve control policies for indoor navigational tasks. We proposed NEAT-GRU which is a modification of NEAT that has the ability to mutate GRUs into the NEAT networks thereby facilitating the use of long term memories. We compared the solutions produced by NEAT and NEAT-GRU to a particular type of bug algorithm called I-Bug which is more suited for use in real robotic domains. We showed that the I-Bug algorithm is not a globally optimal solution and that both NEAT and NEAT-GRU can evolve solutions that outperform it repeatably. We introduced a harder domain in which only the distance to the target of the maze is given as sensor input to the network and demonstrated that NEAT-GRU can successfully evolve solutions to this task, where as NEAT fails in every run. This demonstrates that significant cognition is required to solve this task and that long term memory units greatly assist in these types of tasks found in real robotic domains. 




\bibliographystyle{unsrt}  
\bibliography{main-arxiv.bib}  

\begin{thebibliography}{10}

\bibitem{handbook_robotics}
B~Siciliano and O~Khatib.
\newblock {\em {Springer Handbook of Robotics}}.
\newblock Springer-Verlag, Berlin, Heidelberg, 2007.

\bibitem{bigdog}
D~Wooden, M~Malchano, K~Blankespoor, A~Howardy, A~A Rizzi, and M~Raibert.
\newblock {Autonomous navigation for BigDog}.
\newblock In {\em 2010 IEEE International Conference on Robotics and
  Automation}, pages 4736--4741, may 2010.

\bibitem{range}
A~Bachrach, A~de~Winter, R~He, G~Hemann, S~Prentice, and N~Roy.
\newblock {RANGE - robust autonomous navigation in GPS-denied environments}.
\newblock In {\em 2010 IEEE International Conference on Robotics and
  Automation}, pages 1096--1097, may 2010.

\bibitem{droneslam}
Shaojie Shen and Vijay Kumar.
\newblock {3D Indoor Exploration with a Computationally Constrained MAV}.
\newblock In {\em Robotics: Science and Systems}, 2011.

\bibitem{Thrun2006}
Sebastian Thrun, Mark Diel, and Dirk H{\"{a}}hnel.
\newblock {\em {Scan Alignment and 3-D Surface Modeling with a Helicopter
  Platform}}, pages 287--297.
\newblock Springer Berlin Heidelberg, Berlin, Heidelberg, 2006.

\bibitem{bresson}
G~Bresson, Z~Alsayed, L~Yu, and S~Glaser.
\newblock {Simultaneous Localization and Mapping: A Survey of Current Trends in
  Autonomous Driving}.
\newblock {\em IEEE Transactions on Intelligent Vehicles}, 2(3):194--220, sep
  2017.

\bibitem{Mnih2015}
Volodymyr Mnih, Koray Kavukcuoglu, David Silver, Andrei~A Rusu, Joel Veness,
  Marc~G Bellemare, Alex Graves, Martin Riedmiller, Andreas~K Fidjeland, Georg
  Ostrovski, Stig Petersen, Charles Beattie, Amir Sadik, Ioannis Antonoglou,
  Helen King, Dharshan Kumaran, Daan Wierstra, Shane Legg, and Demis Hassabis.
\newblock {Human-level control through deep reinforcement learning}.
\newblock {\em Nature}, 518:529, feb 2015.

\bibitem{atari_2}
Tobias Pohlen, Bilal Piot, Todd Hester, Mohammad Azar, Dan Horgan, David
  Budden, Gabriel Barth-Maron, Hado {Van Hasselt}, John Quan, Mel
  Ve{\v{c}}er{\'{i}}k, Matteo Hessel, Remi Munos, and Olivier Pietquin.
\newblock {Observe and Look Further: Achieving Consistent Performance on
  Atari}.
\newblock {\em CoRR}, 2018.

\bibitem{DeepNeuro}
Felipe~Petroski Such, Vashisht Madhavan, Edoardo Conti, Joel Lehman, Kenneth~O
  Stanley, and Jeff Clune.
\newblock {Deep Neuroevolution: Genetic Algorithms Are a Competitive
  Alternative for Training Deep Neural Networks for Reinforcement Learning}.
\newblock {\em CoRR}, 2017.

\bibitem{NEAT}
Kenneth~O Stanley and Risto Miikkulainen.
\newblock {Evolving Neural Networks Through Augmenting Topologies}.
\newblock {\em Evolutionary Computation}, 10(2):99--127, 2002.

\bibitem{footbot}
M~Dorigo, D~Floreano, L~M Gambardella, F~Mondada, S~Nolfi, T~Baaboura,
  M~Birattari, M~Bonani, M~Brambilla, A~Brutschy, D~Burnier, A~Campo, A~L
  Christensen, A~Decugniere, G~{Di Caro}, F~Ducatelle, E~Ferrante, A~Forster,
  J~M Gonzales, J~Guzzi, V~Longchamp, S~Magnenat, N~Mathews, M~{Montes de Oca},
  R~O'Grady, C~Pinciroli, G~Pini, P~Retornaz, J~Roberts, V~Sperati, T~Stirling,
  A~Stranieri, T~Stutzle, V~Trianni, E~Tuci, A~E Turgut, and F~Vaussard.
\newblock {Swarmanoid: A Novel Concept for the Study of Heterogeneous Robotic
  Swarms}.
\newblock {\em IEEE Robotics Automation Magazine}, 20(4):60--71, 2013.

\bibitem{gru}
Kyunghyun Cho, Bart van Merrienboer, {\c{C}}aglar G{\"{u}}l{\c{c}}ehre, Fethi
  Bougares, Holger Schwenk, and Yoshua Bengio.
\newblock {Learning Phrase Representations using RNN Encoder-Decoder for
  Statistical Machine Translation}.
\newblock {\em CoRR}, 2014.

\bibitem{com}
V~Lumelsky and A~Stepanov.
\newblock {Dynamic path planning for a mobile automaton with limited
  information on the environment}.
\newblock {\em IEEE Transactions on Automatic Control}, 31(11):1058--1063, nov
  1986.

\bibitem{alg1}
A~Sankaranarayanan and M~Vidyasagar.
\newblock {A new path planning algorithm for moving a point object amidst
  unknown obstacles in a plane}.
\newblock In {\em Proceedings., IEEE International Conference on Robotics and
  Automation}, pages 1930--1936 vol.3, may 1990.

\bibitem{alg2}
A~Sankaranarayanar and M~Vidyasagar.
\newblock {Path planning for moving a point object amidst unknown obstacles in
  a plane: a new algorithm and a general theory for algorithm development}.
\newblock In {\em 29th IEEE Conference on Decision and Control}, pages
  1111--1119 vol.2, dec 1990.

\bibitem{vis1}
V~J Lumelsky and T~Skewis.
\newblock {Incorporating range sensing in the robot navigation function}.
\newblock {\em IEEE Transactions on Systems, Man, and Cybernetics},
  20(5):1058--1069, sep 1990.

\bibitem{vis2}
V~Lumelsky and T~Skewis.
\newblock {A paradigm for incorporating vision in the robot navigation
  function}.
\newblock In {\em Proceedings. 1988 IEEE International Conference on Robotics
  and Automation}, pages 734--739 vol.2, apr 1988.

\bibitem{tanbug}
I~Kamon, E~Rivlin, and E~Rimon.
\newblock {A new range-sensor based globally convergent navigation algorithm
  for mobile robots}.
\newblock In {\em Proceedings of IEEE International Conference on Robotics and
  Automation}, volume~1, pages 429--435 vol.1, apr 1996.

\bibitem{cautious}
E~Magid and E~Rivlin.
\newblock {CautiousBug: a competitive algorithm for sensory-based robot
  navigation}.
\newblock In {\em 2004 IEEE/RSJ International Conference on Intelligent Robots
  and Systems (IROS) (IEEE Cat. No.04CH37566)}, volume~3, pages 2757--2762
  vol.3, sep 2004.

\bibitem{insert}
Qi-Lei Xu and Gong-You Tang.
\newblock {Vectorization path planning for autonomous mobile agent in unknown
  environment}.
\newblock {\em Neural Computing and Applications}, 23(7):2129--2135, dec 2013.

\bibitem{distbug}
I~Kamon and E~Rivlin.
\newblock {Sensory-based motion planning with global proofs}.
\newblock {\em IEEE Transactions on Robotics and Automation}, 13(6):814--822,
  dec 1997.

\bibitem{ibug}
K~Taylor and S~M LaValle.
\newblock {I-Bug: An intensity-based bug algorithm}.
\newblock In {\em 2009 IEEE International Conference on Robotics and
  Automation}, pages 3981--3986, may 2009.

\bibitem{basti}
Bastian Broecker, Karl Tuyls, and James Butterworth.
\newblock {Distance-based multi-robot coordination on pocket drones}.
\newblock In {\em IEEE International Conference on Robotics and Automation
  (ICRA)}, Brisbane, 2018.

\bibitem{Shorten2015}
D~Shorten and G~Nitschke.
\newblock {Evolving Generalised Maze Solvers}.
\newblock In Antonio~M Mora and Giovanni Squillero, editors, {\em Applications
  of Evolutionary Computation}, pages 783--794, Cham, 2015. Springer
  International Publishing.

\bibitem{NS1}
Joel Lehman and Kenneth~O Stanley.
\newblock {Exploiting open-endedness to solve problems through the search for
  novelty}.
\newblock In {\em Proceedings of the Eleventh International Conference on
  Artificial Life (Alife XI)}. MIT Press, 2008.

\bibitem{NS3}
Joel Lehman and Kenneth~O Stanley.
\newblock {Abandoning Objectives: Evolution Through the Search for Novelty
  Alone}.
\newblock {\em Evol. Comput.}, 19(2):189--223, jun 2011.

\bibitem{NS2}
Joel Lehman and Kenneth~O Stanley.
\newblock {\em {Novelty Search and the Problem with Objectives}}, pages 37--56.
\newblock Springer New York, New York, NY, 2011.

\bibitem{QD}
Justin~K Pugh, Lisa~B Soros, and Kenneth~O Stanley.
\newblock {Quality Diversity: A New Frontier for Evolutionary Computation}.
\newblock {\em Frontiers in Robotics and AI}, 3:40, 2016.

\bibitem{Soltoggio2008}
Andrea Soltoggio, John Bullinaria, Claudio Mattiussi, Peter D{\"{u}}rr, and
  Dario Floreano.
\newblock {Evolutionary Advantages of Neuromodulated Plasticity in Dynamic,
  Reward- based Scenarios}.
\newblock {\em ALIFE}, 2008.

\bibitem{Risi2009}
Sebastian Risi, Sandy~D Vanderbleek, Charles~E Hughes, and Kenneth~O Stanley.
\newblock {How Novelty Search Escapes the Deceptive Trap of Learning to Learn}.
\newblock In {\em Proceedings of the 11th Annual Conference on Genetic and
  Evolutionary Computation}, GECCO '09, pages 153--160, New York, NY, USA,
  2009. ACM.

\bibitem{Risi2010NS}
Sebastian Risi, Charles~E Hughes, and Kenneth~O Stanley.
\newblock {Evolving Plastic Neural Networks with Novelty Search}.
\newblock {\em Adaptive Behavior - Animals, Animats, Software Agents, Robots,
  Adaptive Systems}, 18(6):470--491, dec 2010.

\bibitem{Ollion2012}
Charles Ollion, Tony Pinville, and St{\'{e}}phane Doncieux.
\newblock {Emergence of memory in neuroevolution: Impact of selection
  pressures}.
\newblock {\em GECCO'12 - Proceedings of the 14th International Conference on
  Genetic and Evolutionary Computation Companion}, 2012.

\bibitem{Gomez2005}
Faustino~J Gomez and J{\"{u}}rgen Schmidhuber.
\newblock {Co-evolving Recurrent Neurons Learn Deep Memory POMDPs}.
\newblock In {\em Proceedings of the 7th Annual Conference on Genetic and
  Evolutionary Computation}, GECCO '05, pages 491--498, New York, NY, USA,
  2005. ACM.

\bibitem{Risi2010}
Sebastian Risi and Kenneth~O Stanley.
\newblock {Indirectly Encoding Neural Plasticity As a Pattern of Local Rules}.
\newblock In {\em Proceedings of the 11th International Conference on
  Simulation of Adaptive Behavior: From Animals to Animats}, SAB'10, pages
  533--543, Berlin, Heidelberg, 2010. Springer-Verlag.

\bibitem{Risi2012}
S~Risi and K~O Stanley.
\newblock {A unified approach to evolving plasticity and neural geometry}.
\newblock In {\em The 2012 International Joint Conference on Neural Networks
  (IJCNN)}, pages 1--8, jun 2012.

\bibitem{HyperNEAT}
Kenneth~O Stanley, David~B D'Ambrosio, and Jason Gauci.
\newblock {A Hypercube-based Encoding for Evolving Large-scale Neural
  Networks}.
\newblock {\em Artif. Life}, 15(2):185--212, apr 2009.

\bibitem{ENTM1}
Rasmus~Boll Greve, Emil~Juul Jacobsen, and Sebastian Risi.
\newblock {Evolving Neural Turing Machines for Reward-based Learning}.
\newblock In {\em Proceedings of the Genetic and Evolutionary Computation
  Conference 2016}, GECCO '16, pages 117--124, New York, NY, USA, 2016. ACM.

\bibitem{ENTM2}
Benno L{\"{u}}ders, Mikkel Schl{\"{a}}ger, and Sebastian Risi.
\newblock {Continual Learning through Evolvable Neural Turing Machines}.
\newblock In {\em Proceedings of the NIPS 2016 Workshop on Continual Learning
  and Deep Networks (CLDL 2016)}, 2016.

\bibitem{ENTM3}
Benno L{\"{u}}ders, Mikkel Schl{\"{a}}ger, Aleksandra Korach, and Sebastian
  Risi.
\newblock {Continual and One-Shot Learning Through Neural Networks with Dynamic
  External Memory}.
\newblock In {\em EvoApplications}, 2017.

\bibitem{GRUMB}
Shauharda Khadka, Jen {Jen Chung}, and Kagan Tumer.
\newblock {Evolving memory-augmented neural architecture for deep memory
  problems}.
\newblock In {\em Proceedings of the Genetic and Evolutionary Computation
  Conference 2017}, pages 441--448, 2017.

\bibitem{CTRNN}
Jesper Blynel and Dario Floreano.
\newblock {Exploring the T-Maze: Evolving Learning-Like Robot Behaviors Using
  CTRNNs}.
\newblock In {\em Applications of Evolutionary Computing}, pages 593--604,
  Berlin, Heidelberg, 2003. Springer Berlin Heidelberg.

\bibitem{MCNS}
Joel Lehman and Kenneth~O Stanley.
\newblock {Revising the Evolutionary Computation Abstraction: Minimal Criteria
  Novelty Search}.
\newblock In {\em Proceedings of the 12th Annual Conference on Genetic and
  Evolutionary Computation}, GECCO '10, pages 103--110, New York, NY, USA,
  2010. ACM.

\bibitem{MCC}
Jonathan {C. Brant} and Kenneth {O. Stanley}.
\newblock {Minimal criterion coevolution: a new approach to open-ended search}.
\newblock In {\em Proceedings of the Genetic and Evolutionary Computation
  Conference 2017}, pages 67--74, 2017.

\bibitem{Tessler}
Chen Tessler, Shahar Givony, Tom Zahavy, Daniel Mankowitz, and Shie Mannor.
\newblock {A Deep Hierarchical Approach to Lifelong Learning in Minecraft}.
\newblock In {\em CoRR}, 2016.

\bibitem{Kempka}
Michal Kempka, Marek Wydmuch, Grzegorz Runc, Jakub Toczek, and Wojciech
  Jaskowski.
\newblock {ViZDoom: A Doom-based AI Research Platform for Visual Reinforcement
  Learning}.
\newblock {\em CoRR}, 2016.

\bibitem{Kulkarni}
Tejas {D. Kulkarni}, Ardavan Saeedi, Simanta Gautam, and Samuel {J. Gershman}.
\newblock {Deep Successor Reinforcement Learning}.
\newblock {\em CoRR}, 2016.

\bibitem{robot_3}
Yuke Zhu, Roozbeh Mottaghi, Eric Kolve, Joseph {J. Lim}, Abhinav Gupta,
  Li~Fei-Fei, and Ali Farhadi.
\newblock {Target-driven Visual Navigation in Indoor Scenes using Deep
  Reinforcement Learning}.
\newblock {\em CoRR}, 2016.

\bibitem{Tai}
Lei Tai and Ming Liu.
\newblock {Towards Cognitive Exploration through Deep Reinforcement Learning
  for Mobile Robots}.
\newblock {\em CoRR}, 2016.

\bibitem{Mirowski}
Piotr~W Mirowski, Razvan Pascanu, Fabio Viola, Hubert Soyer, Andrew~J Ballard,
  Andrea Banino, Misha Denil, Ross Goroshin, Laurent Sifre, Koray Kavukcuoglu,
  Dharshan Kumaran, and Raia Hadsell.
\newblock {Learning to Navigate in Complex Environments}.
\newblock {\em CoRR}, 2016.

\bibitem{Jaderberg}
Max Jaderberg, Volodymyr Mnih, Wojciech~Marian Czarnecki, Tom Schaul, Joel~Z
  Leibo, David Silver, and Koray Kavukcuoglu.
\newblock {Reinforcement Learning with Unsupervised Auxiliary Tasks}.
\newblock {\em CoRR}, abs/1611.0, 2016.

\bibitem{Mnih}
Volodymyr Mnih, Adria~Puigdomenech Badia, Mehdi Mirza, Alex Graves, Timothy
  Lillicrap, Tim Harley, David Silver, and Koray Kavukcuoglu.
\newblock {Asynchronous Methods for Deep Reinforcement Learning}.
\newblock In Maria~Florina Balcan and Kilian~Q Weinberger, editors, {\em
  Proceedings of The 33rd International Conference on Machine Learning},
  volume~48 of {\em Proceedings of Machine Learning Research}, pages
  1928--1937, New York, New York, USA, 2016. PMLR.

\bibitem{Oh}
Junhyuk Oh, Valliappa Chockalingam, Satinder Singh, and Honglak Lee.
\newblock {Control of Memory, Active Perception, and Action in Minecraft}.
\newblock In {\em Proceedings of the 33rd International Conference on
  International Conference on Machine Learning - Volume 48}, ICML'16, pages
  2790--2799. JMLR.org, 2016.

\bibitem{Bakker}
Bram Bakker.
\newblock {Reinforcement Learning with Long Short-term Memory}.
\newblock In {\em Proceedings of the 14th International Conference on Neural
  Information Processing Systems: Natural and Synthetic}, NIPS'01, pages
  1475--1482, Cambridge, MA, USA, 2001. MIT Press.

\bibitem{Pfeiffer}
Mark Pfeiffer, Samarth Shukla, Matteo Turchetta, Cesar Cadena, Andreas Krause,
  Roland Siegwart, and Juan~I Nieto.
\newblock {Reinforced Imitation: Sample Efficient Deep Reinforcement Learning
  for Map-less Navigation by Leveraging Prior Demonstrations}.
\newblock {\em CoRR}, 2018.

\bibitem{grucompare}
Junyoung Chung, {\c{C}}aglar G{\"{u}}l{\c{c}}ehre, KyungHyun Cho, and Yoshua
  Bengio.
\newblock {Empirical Evaluation of Gated Recurrent Neural Networks on Sequence
  Modeling}.
\newblock {\em CoRR}, 2014.

\bibitem{rawal}
A~Rawal and R~Miikkulainen.
\newblock {Evolving Deep LSTM-based Memory Networks Using an Information
  Maximization Objective}.
\newblock In {\em GECCO 2016}, GECCO '16, pages 501--508, 2016.

\bibitem{argos}
C~Pinciroli, V~Trianni, R~O'Grady, G~I Pini, A~Brutschy, M~Brambilla,
  N~Mathews, E~Ferrante, G~{Di Caro}, F~Ducatelle, M~Birattari, L~M
  Gambardella, and M~Dorigo.
\newblock {ARGoS: A Modular, Parallel, Multi-Engine Simulator for Multi-Robot
  Systems}.
\newblock {\em Swarm Intelligence}, 6(4):271--295, 2012.

\bibitem{kim_bug}
K~McGuire, G~de~Croon, and K~Tuyls.
\newblock {A Comparative Study of Bug Algorithms for Robot Navigation}.
\newblock {\em CoRR}, 2018.

\end{thebibliography}


\end{document}